\documentclass[runningheads]{llncs}
\usepackage{graphicx}
\usepackage[colorlinks]{hyperref}
\usepackage{cite}
\usepackage{amsmath,amssymb,amsfonts}
\usepackage{textcomp}
\usepackage{xcolor}
\usepackage{algpseudocode}
\usepackage{amsmath}
\usepackage{algorithm}
\usepackage{graphics}
\usepackage{caption}
\usepackage{subcaption}
\usepackage{tabularx}
\usepackage{flushend}
\usepackage[normalem]{ulem}
\usepackage[bottom]{footmisc}

\date{}

\usepackage[overlay,absolute]{textpos}

\begin{document}

\begin{textblock*}{10in}(38mm, 10mm)
{\textbf{Ref:} \emph{International Conference on Artificial Neural Networks (ICANN)}, Springer LNCS,}
\end{textblock*}
\begin{textblock*}{10in}(38mm, 15mm)
{Vol.~11731, pp.~440--455, Munich, Germany, September 2019.}
\end{textblock*}

\title{DeepMimic: Mentor-Student Unlabeled\\
Data Based Training}

\author{Itay Mosafi \and Eli (Omid) David \and Nathan S. Netanyahu}

\authorrunning{I.~Mosafi, E.O.~David, and N.S.~Netanyahu}

\institute{
Department of Computer Science, Bar-Ilan University, Ramat-Gan 5290002, Israel \\
\email{itay.mosafi@gmail.com, mail@elidavid.com, nathan@cs.biu.ac.il}
}

\makeatletter
\def\thickhline{%
  \noalign{\ifnum0=`}\fi\hrule \@height \thickarrayrulewidth \futurelet
   \reserved@a\@xthickhline}
\def\@xthickhline{\ifx\reserved@a\thickhline
               \vskip\doublerulesep
               \vskip-\thickarrayrulewidth
             \fi
      \ifnum0=`{\fi}}
\makeatother

\newlength{\thickarrayrulewidth}
\setlength{\thickarrayrulewidth}{2\arrayrulewidth}

\newcommand\setrow[1]{\gdef\rowmac{#1}#1\ignorespaces}
\newcommand\clearrow{\global\let\rowmac\relax}
\clearrow

\maketitle
\begin{abstract}
In this paper, we present a deep neural network (DNN) training approach called the ``DeepMimic'' training method. Enormous amounts of data are available nowadays for training usage. Yet, only a tiny portion of these data is manually labeled, whereas almost all of the data are unlabeled.
The training approach presented utilizes, in a most simplified manner, the unlabeled data to the fullest, in order to achieve remarkable (classification) results.
Our DeepMimic method uses a small portion of labeled data and a large amount of unlabeled data for the training process, as expected in a real-world scenario. It consists of a mentor model and a student model.
Employing a mentor model trained on a small portion of the labeled data and then feeding it only with unlabeled data, we show how to obtain a (simplified) student model that reaches the same accuracy and loss as the mentor model, on the same test set, without using any of the original data labels in the training of the student model.
Our experiments demonstrate that even on challenging classification tasks the student network architecture can be simplified significantly with a minor influence on the performance, i.e., we need not even know the original network architecture of the mentor. In addition, the time required for training the student model to reach the mentor's performance level is shorter, as a result of a simplified architecture and more available data. 
The proposed method highlights the disadvantages of regular supervised training and demonstrates the benefits of a less traditional training approach.
\end{abstract}
\section{Introduction}
Deep neural networks (DNNs) have been used lately very effectively in many applications, e.g., object detection (as described by the ImageNet challenge \cite{imagenet2009}), with state-of-the-art performance \cite{DBLP:journals/corr/abs-1709-01507} exceeding human-level capabilities, natural language processing, where text translation using DNNs with attention mechanism \cite{LuongPM15} has achieved remarkable results, playing highly-complex games (such as chess \cite{david2016deepchess} and Go \cite{silver2017mastering}) at a grandmaster level, generation of realistic-looking images \cite{goodfellow2014generative}, etc.

The recent impressive advancement of deep learning (DL) can be attributed to a number of factors, including: (1) Enhancement of computational capabilities (e.g., using strong graphical processing units (GPUs)), (2) improvement of network architectures, and (3) acquisition of vast amounts of training data. 
With the growing availability of powerful computational capabilities, much of the research has focused on innovative network architectures for the pursuit of state-of-the-art performance in various problem domains. Some examples include: {\it transferable architectures} \cite{DBLP:journals/corr/ZophVSL17}, which suggest a method of learning the model architectures directly on the dataset of interest, {\it fractional max-pooling} \cite{DBLP:journals/corr/Graham14a}, which offers a modification to the standard max-pooling in convolutional neural networks (CNNs) \cite{cnn1998}, and {\it exponential linear units} (ELUs) \cite{ClevertUH15}, which provide a new activation function for improving learning characteristics.

In this paper, we focus mainly on the usage of large amounts of available unlabeled data to form a training method that utilizes these available resources.
Specifically, we focus here on a new DeepMimic training methodology, demonstrating its effectiveness with respect to object classification based on the use of CNNs.

Occupied mainly by the performance of DNNs in numerous applications, researchers may tend to overlook various aspects of the learning process, e.g., the specific manner in which supervised learning (i.e., the training of a network using labeled data) is performed.
In the case of multi-label classification, each data item is associated with a class label, and is represented by a ``one-hot'' encoding vector. (The dimension of a one-hot encoding vector is the number of possible classes in the dataset, such that, the correct class index contains '1' and all the other indexes contain '0'.)
It is reasonable to assume that a label distribution that is different from the one-hot vector representation might gain extra insight or knowledge about the model, thereby changing significantly the training process.

To explore this idea, we need a meaningful label distribution, which we gain by using the proposed DeepMimic paradigm. In our method, we use a relatively small subset of our data to perform supervised training on our mentor model, while treating the rest of the dataset as unlabeled data, i.e., ignoring the labels completely. Once the mentor is trained, we use it to create a label distribution by outputting the softmax components for each data item in the unlabeled dataset.
During the data splitting process, the one-hot labels are used merely to ensure a balanced dataset split. We later show that this might not be actually required, based on our empirical results for the unbalanced dataset, which yield the same accuracy gained for the balanced dataset.

Using the unlabeled data and the label distribution produced by the mentor model we train a student model. We are able to achieve comparable performance to the mentor's, with a student model that is simpler, shallower, and substantially faster.
In other words, our method can extract a model's knowledge and successfully transfer it to another model using essentially no labeled data.
These remarkable results suggest that the method presented can be used in many applications. One can take advantage of large amounts of unlabeled data and mimic a black-box trained model, without even knowing its architecture or the labeled data used for its training. For example, an individual can purchase a neural network-based product and create a copy of it, which will match the original product's performance with no access to the data used to train the product. Finally, the student model could result in a substantially simpler architecture. Therefore, we can achieve a much faster inference time, which is very important in production for various real life systems and services.

\section{Background}
Many real-life problems have led to interesting, innovative DL techniques, such as those pertaining to the 
%ease
mentor-student learning process~\cite{mentor_student_generations}. These methods suggest a less strict training of the mentor with the overall gain of lowering the risk of overfitting by the student. In~\cite{kim2017transferring} a class-distance loss is presented that assists mentor networks in forming densely-clustered vector spaces to make it easy for a student network to learn from. In~\cite{robust_student} the authors focus on enhancing the robustness of the student network without sacrificing the performance. Model compression, originally researched in~\cite{model_compression}, presents a way of compressing the function learned by a complex model into a much smaller, faster model.

The problems addressed in this paper are considered nowadays rather simple, and thus the method should be reestablished on more challenging problems. Furthermore, years ago, when the Internet was much less developed and considerably smaller amounts of data were available, the focus was directed at the ability to generate synthetic data for training and development purposes. With tens of zettabytes ($1000^7 = 10^{21}$ bytes) of data available online, acquiring unlabeled data is no longer an issue. Currently, the main interest is to develop ways of exploiting these data efficiently.

The ability to distill the knowledge of a neural network to create a simpler and more suitable production network \cite{hinton2015distilling}, \cite{do_deep}, \cite{copycat_cnn}, \cite{stealing_knowledge_2019} is extremely valuable. 
During this knowledge transfer, the method of training on soft labels, i.e., using a vector of classes (whose probabilities sum up to 1) as labels, seems to provide much more information for the training process compared to the training with one-hot vectors only. This supports the notion that training based on one-hot labels may not be ideal.

Another interesting aspect of soft-label training is its use of regularization~\cite{aghajanyan2017softtarget}.
Regularization techniques for preventing overfitting and achieving better generalization consist mainly of {\it dropout} \cite{hinton2012improving}, \cite{srivastava2014dropout}, i.e., randomly ``shutting down'' some of the neurons, {\it DropConnect} \cite{wan2013regularization}, for random cancellation of synapses between neurons in a very similar way to dropout, random noise addition~\cite{li2016whiteout}, and weight decay \cite{krogh1992simple}. These techniques are also referred to as $L_1$ and $L_2$ regularization \cite{ng2004feature}.
Another work is the mixup paper \cite{mixup2018}, which shows that averaging the training examples and their labels, e.g., creating a new image and its label as a weighted average of the original two images and two one-hot vectors used as labels,
to improve the regularization.
It is also possible to transfer knowledge from different types of networks, e.g., a recurrent neural network (RNN) to a DNN, as shown in~\cite{chan2015transferring}.

Mimicking a model's predictions in order to obtain knowledge has been researched in various aspects. In~\cite{2016mimic} it is used to transfer knowledge from one domain to another, in order to generalize it and teach a reinforcement learning agent how to behave in multiple tasks simultaneously. In our case, we mimic a mentor model and try to acquire its knowledge as well; yet, we always remain in the same domain and try to maximize the student's performance there.
In~\cite{rusu2015policy} the authors show that their method can extract the policy of a reinforcement learning agent, and train 
a new network, which is dramatically smaller and more efficient, while performing at a comparable level of the agent's.
Thinner and deeper student models are presented in~\cite{romero2015fitnets}; the method discussed allows using not only the outputs but the intermediate representations
learned by the mentor as hints to improve the training process and the final performance of the student. In \cite{urban2017deep} it is argued that even though a student model does not have to be as deep as its mentor, it requires the same number of convolutional layers in order to learn
functions of comparable accuracy.
According to their results, the large gap between CNNs and fully-connected DNNs cannot be significantly reduced, as long as the student model does not consist of multiple convolutional layers.

The difference from our work is that both mentor and student models are trained over the entire dataset, i.e., there are no unique data seen only by the student, as in our case. For now, the state-of-the-art results on any visual tasks are achieved by CNNs, as the classical fully-connected DNNs simply cannot compete with it. Even though the DNN limits can be pushed further \cite{lin2015far}, they are no match for the CNN architecture which relies on local correlations in a given image.
Our method may enable DNNs to overcome this boundary, since the regular training procedures which failed to do so are not used by our method. Note that we can alter the mentor model as we deem fit, and rely on the soft labels it predicts, in order to train a student model, regardless of its architecture.

\section{DeepMimic Training}
\subsection{Data Split}
When it comes to available data, our goal is to simulate real-life scenarios.
In such a case, we would usually have huge amounts of unlabeled data; these data are considered useless, most of the time, unless used for training autoencoders \cite{baldi2012autoencoders}, for example.

In order to simulate such a scenario, we choose a ratio between the mentor's training data and that of the student's, such that there is a sufficient amount of unlabeled data to train the student and a sufficient amount of training data for the mentor model to reach good performance on the test set.
We performed this experiment on the following datasets: MNIST \cite{mnisthandwrittendigit}, CIFAR-10 \cite{cifar10}, and Tiny ImageNet\cite{additional_tiny_imagenetnet}.
All the training data chosen for the student model are treated as unlabeled data, i.e., we ignore the labels as described in the next section.

The ratio chosen for the data split is 1:4, which produced the best results after testing various split ratios, and considering the need for sufficient training data for the student model. All the images are randomly assigned to create balanced datasets in most experiments.
In other words, by splitting the data randomly, we ensure that for each image of a certain label in the mentor dataset, there are four images of the same label in the student training set. 
This way both the mentor and the student datasets contain an equal number of images of each label, i.e.,  the image amount per each class is balanced. In order to simulate scenarios where the available data distribution is unknown and unbalanced, we modified in some of the experiments the student dataset by forcing, e.g., a different number of samples in each class, as described in Section~\ref{cifar10_unbalanced_section}. We did that by either removing a random number of samples from each class in the student dataset or adding a random amount of out-of-domain images to the student training set. Regardless, it seems the student is only bound to its mentor accuracy rate, i.e., even if there were huge amounts of data for the student, it could not be significantly better than its mentor. 
As for testing, we used the original test set of each dataset, respectively, to test both models.
Since the datasets are fairly limited in size, we decided not to split the data to training, validation and testing; instead, we use all of the available data for training and testing.

\begin{figure}[t]
\centering
\includegraphics[width=1\linewidth]{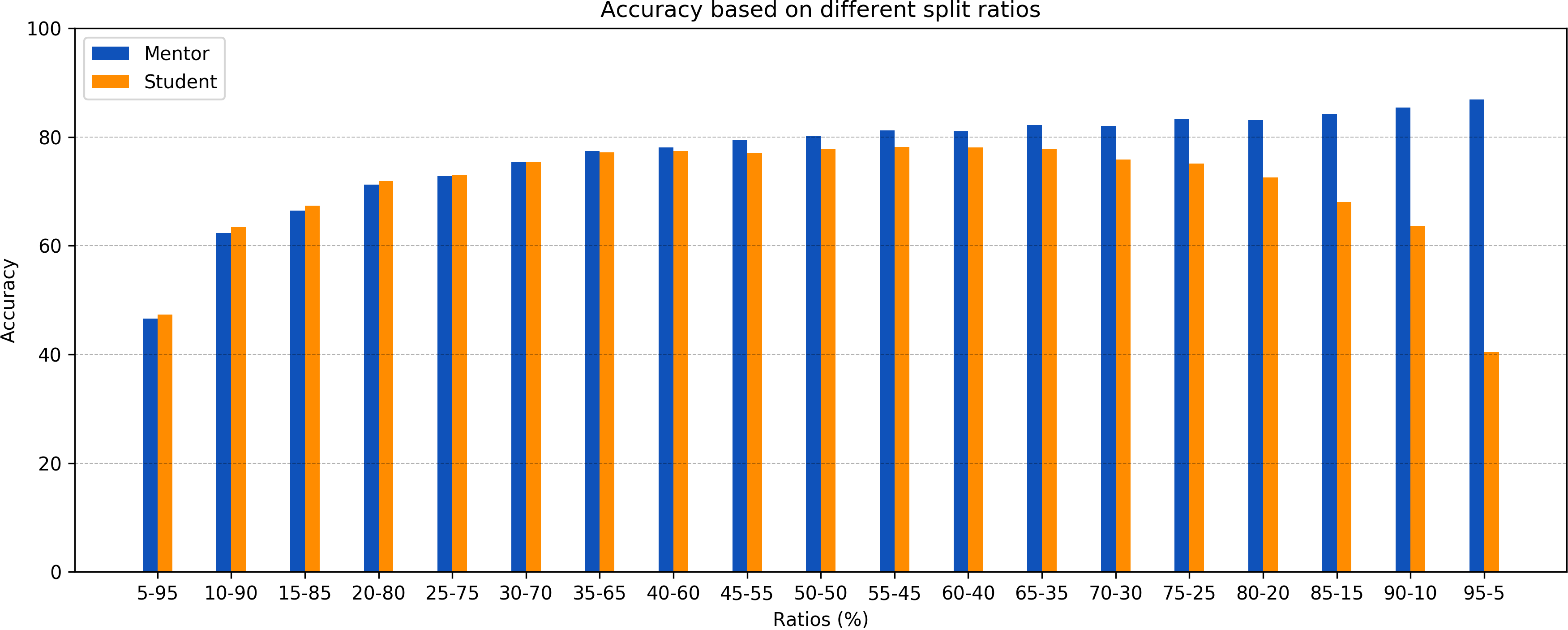}
\label{sfig:testa}
\caption{Accuracy of trained mentor and student on  CIFAR-10 test set as a function of dataset split ratio. Plot depicts the influence of different split ratios on the performance of mentor and student. To allocate sufficient training data to the student and still enable the mentor to reach high accuracy, we choose a 20-80 split (used in the experiments reported for the three different datasets). Presumably, similar ratios to 20-80 with smaller student datasets could work as well, but for more complicated problems, a larger dataset available for the student would probably be required;
this insight could serve as a rule of thumb for desired ratios between labeled and unlabeled data.}
\label{cifar10_accuracy_ratios_bar}
\end{figure}

\subsection{Training Method}
In the training process, we first start by training the mentor model using its assigned dataset. Regularization methods, such as dropout, were vastly used in order to reach high accuracy on the test set.
Considering mainly classification problems, the last layer of each model is a softmax layer, which normalizes the output and provides a distribution for each possible class (with all distributions summing up to 1).
The training uses a stochastic gradient descent (SGD) algorithm and the cross-entropy loss.
Once the mentor is well trained, we can predict a soft label for each image in the student dataset. By doing that, we generate an estimate for each image while still ignoring all the real labels.
We now train the student model, using its assigned data and the soft labels generated by the mentor model.
For the student training, we also use SGD and cross entropy loss.
In the student model training, regularization is less needed since training on the soft labels creates a very strong generalization in the training process \cite{aghajanyan2017softtarget}.
The student reaches the mentor's accuracy on the test set, in all experiments.
Based of the performances of shallow students on test sets, it is clear that the student architecture does not have to be similar to the mentor's, while the performance remains almost identical on the test set. In all the classification tasks we worked on, the reduced student network consistently maintained the mentor's performance.

\section{Experiments}
\subsection{MNIST}

\begin{table}[b]
\begin{center}
\def\arraystretch{1.5}%
\begin{tabular}{cccc}
\thickhline
\setrow{\bfseries}Model &\setrow{\bfseries} Architecture &\setrow{\bfseries} Accuracy &\setrow{\bfseries} Relative  Accuracy\\
\thickhline
    Mentor & $c-mp-c-mp-fc^2-s$ & 97.46\% & -\\
    Student-A & $c-mp-c-mp-fc^2-s$ & 97.38\%& 99.91\%\\
    Student-B & $c-mp-fc^2-s$ & 97.17\% & 99.70\%\\
	\thickhline
\end{tabular}
\caption{Model architectures, test accuracy, and relative accuracy between Students and Mentor for MNIST dataset. Symbols: c-convolutional layer, mp-max pooling layer, fc-fully connected layer, s-softmax layer. $\theta^n$ means $n$ consecutive layers of type $\theta$.}
\label{mnist_table}
\end{center}
\end{table}

MNIST is a relatively simple dataset containing handwritten digit images; it is ideal to perform a ``sanity check'' on the method. It contains 70,000 ($28 \times 28$) grayscale images, 60,000 of which for training the model and the remaining 10,000 for testing it.

\begin{figure}[t]
\centering
\begin{subfigure}{.5\linewidth} %\hfill% or \hspace{5mm} or \hspace{0.3\textwidth}
\includegraphics[width=.99\linewidth]{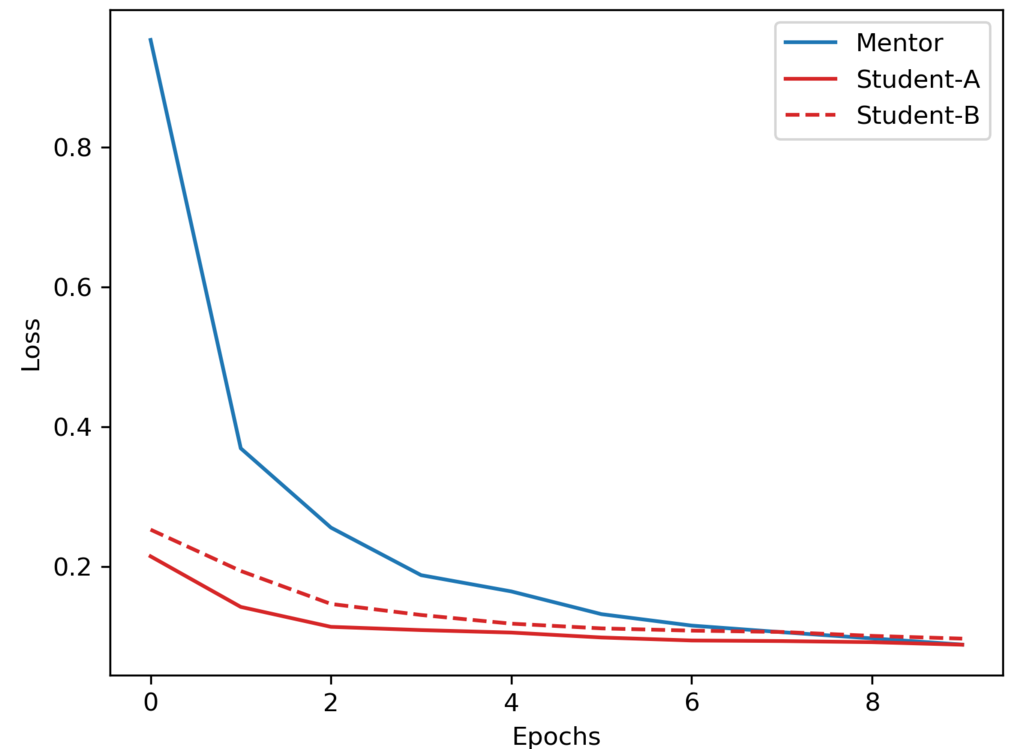}
\caption{Test Loss}
\label{sfig:testa}
\end{subfigure}%
\begin{subfigure}{.5\linewidth}
\includegraphics[width=.99\linewidth]{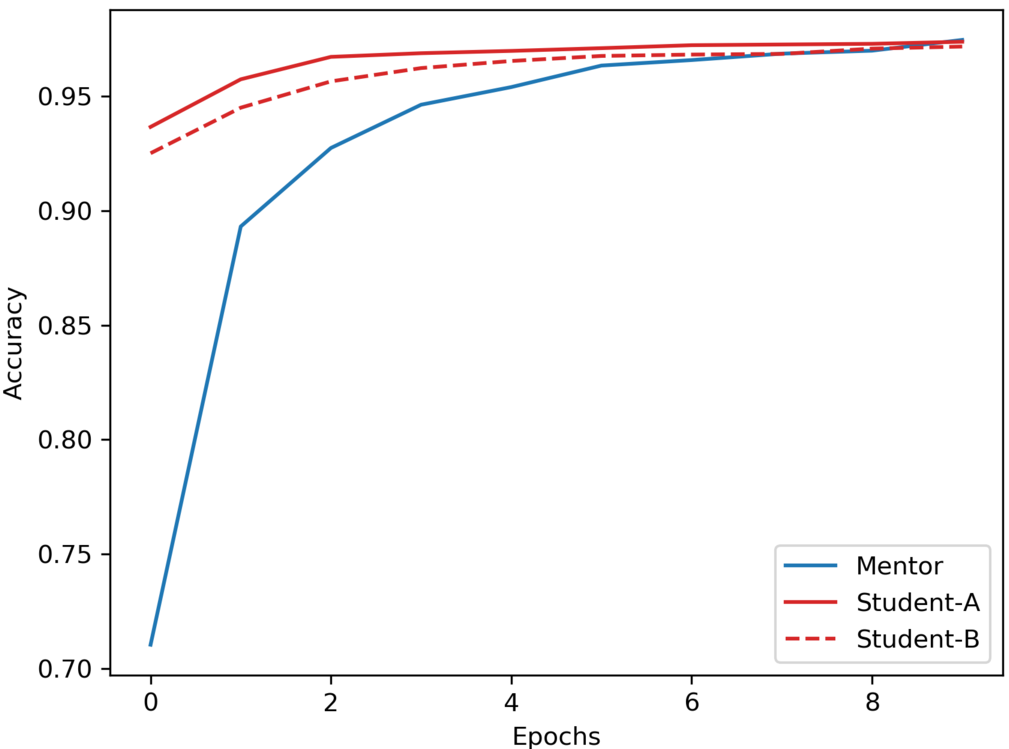}
\caption{Test Accuracy}
\label{sfig:testb}
\end{subfigure}%
\caption{Models' test loss and accuracy for MNIST dataset; even shallower Student-B model reaches almost identical results to the Mentor's.}
\label{mnist_acc_loss_fig}
\end{figure}

As mentioned in the previous section, we use 20\% of the training set for the mentor training; after it is trained, we
use the remaining 80\% and the trained mentor model to create the soft label distributions.
In the experiments reported below, we tested a student model identical to the mentor model, as well as shallower and more simplified student models.
The mentor's accuracy is relative to the amount of data used for training; it is not expected to reach state-of-the-art results with only one fifth of the original training data. This is true for all models trained on a small subset of the standard dataset. As can be seen from Table~\ref{mnist_table} and Figure~\ref{mnist_acc_loss_fig}, the Mentor and Student-A (i.e., the model with the identical architecture) reach almost identical results (i.e., identical loss and accuracy) on the test set, while all the unlabeled data used for training Student-A are never used to train the Mentor. Student-B reaches very close results, as well, i.e., it is possible to create a rather simplified student model to mimic successfully a mentor without knowing its architecture.

\subsection{CIFAR-10}
\begin{table}[!h]
\begin{center}
\def\arraystretch{1.3}%
\begin{tabular}{cccc}
\thickhline
\setrow{\bfseries}Model &\setrow{\bfseries} Architecture &\setrow{\bfseries} Accuracy &\setrow{\bfseries} Relative  Accuracy\\
\thickhline
    Mentor & $c^2-mp-c^2-mp-c^2-mp-fc^2-s$ & 73.14\% & -\\
    Student-A & $c^2-mp-c^2-mp-c^2-mp-fc^2-s$ & 73.58\%& 100.6\%\\
    Student-B & $c^2-mp-c-fc^2-s$ & 72.38\% & 98.96\%\\
    Student-C & $c^2-mp-fc^2-s$ & 69.63\% & 95.2\%\\
	\thickhline
\end{tabular}
\caption{Model architectures, test accuracy, and relative accuracy between Students and Mentor for CIFAR-10 dataset. Symbols: c-convolutional layer, mp-max pooling layer, fc-fully connected layer, s-softmax layer. $\theta^n$ means $n$ consecutive layers of type $\theta$.}
\label{cifar_table}
\end{center}
\end{table}
CIFAR-10 is an established dataset used for object recognition. It consists of 60,000 ($32 \times 32$) RGB images in 10 classes, with 6,000 images per class. There are 50,000 training images and 10,000 test images in the official data.
We used deeper networks for this task; as before, the student networks manage to achieve very good results compared to the mentor's, using various network architectures.\newline

\begin{figure}[!h]
  \centering
  %\hspace*{-0.2cm}
  \includegraphics[width=1\linewidth]{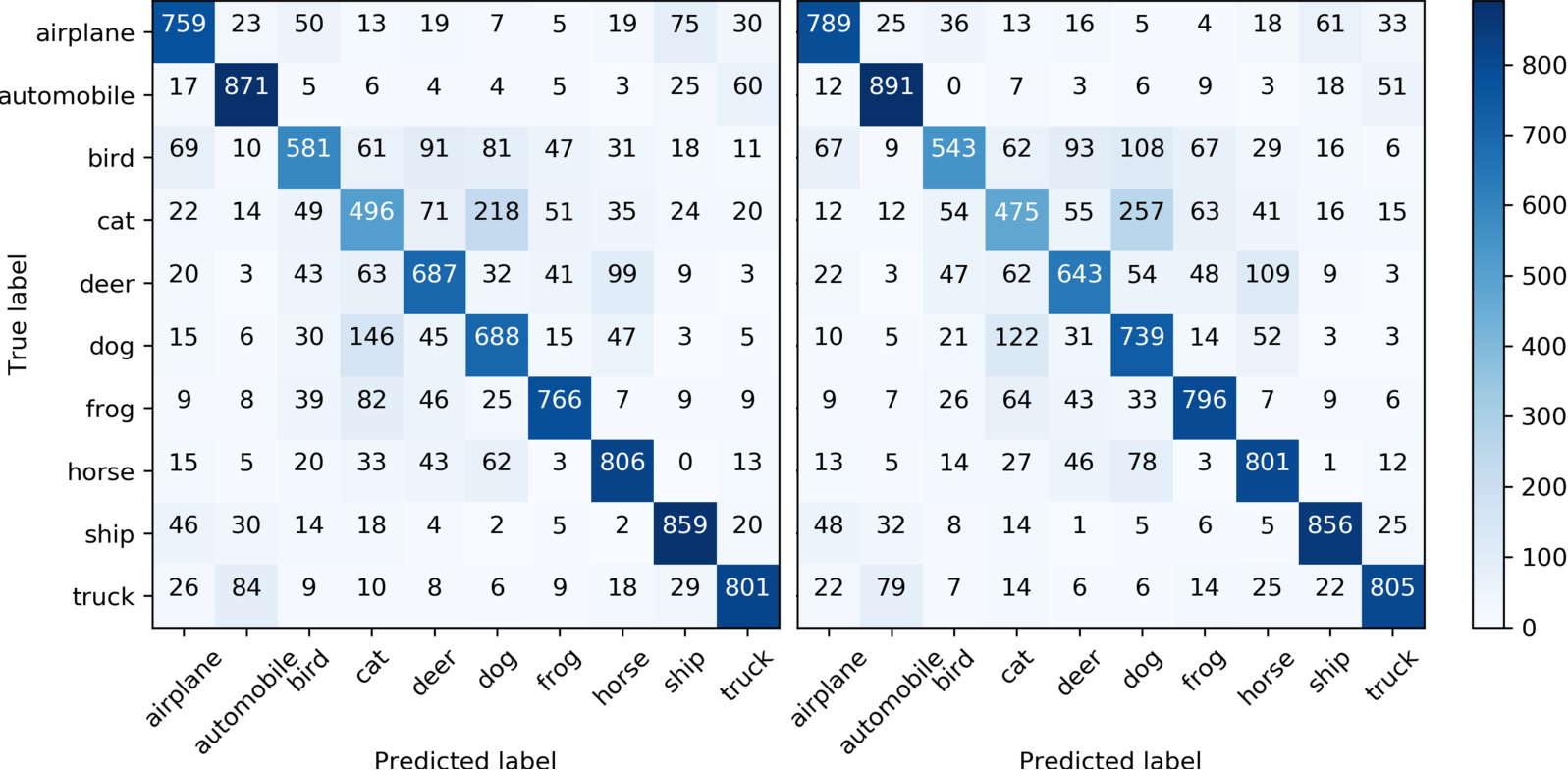}%{.25\linewidth}
  \caption{Confusion matrices of Mentor (LHS) and Student-A (RHS) for CIFAR-10 dataset. Note model's frequency of true and false class predictions for each and every class in test set. This helps understand the degree of confusion in the model, with respect to certain classes, e.g., the model sometimes mistakes a Dog for a Cat and vice versa. A confusion matrix is much more informative than an accuracy measurement. Note that where Mentor tends to make mistakes, so does Student, i.e., they are very similar in all aspects. This best illustrates the successful knowledge transfer from Mentor to Student.}
  \label{fig:confusion_matrices}
\end{figure}

\begin{figure}[!h]
\centering
\begin{subfigure}{.5\linewidth} %\hfill% or \hspace{5mm} or \hspace{0.3\textwidth}
\includegraphics[width=.99\linewidth]{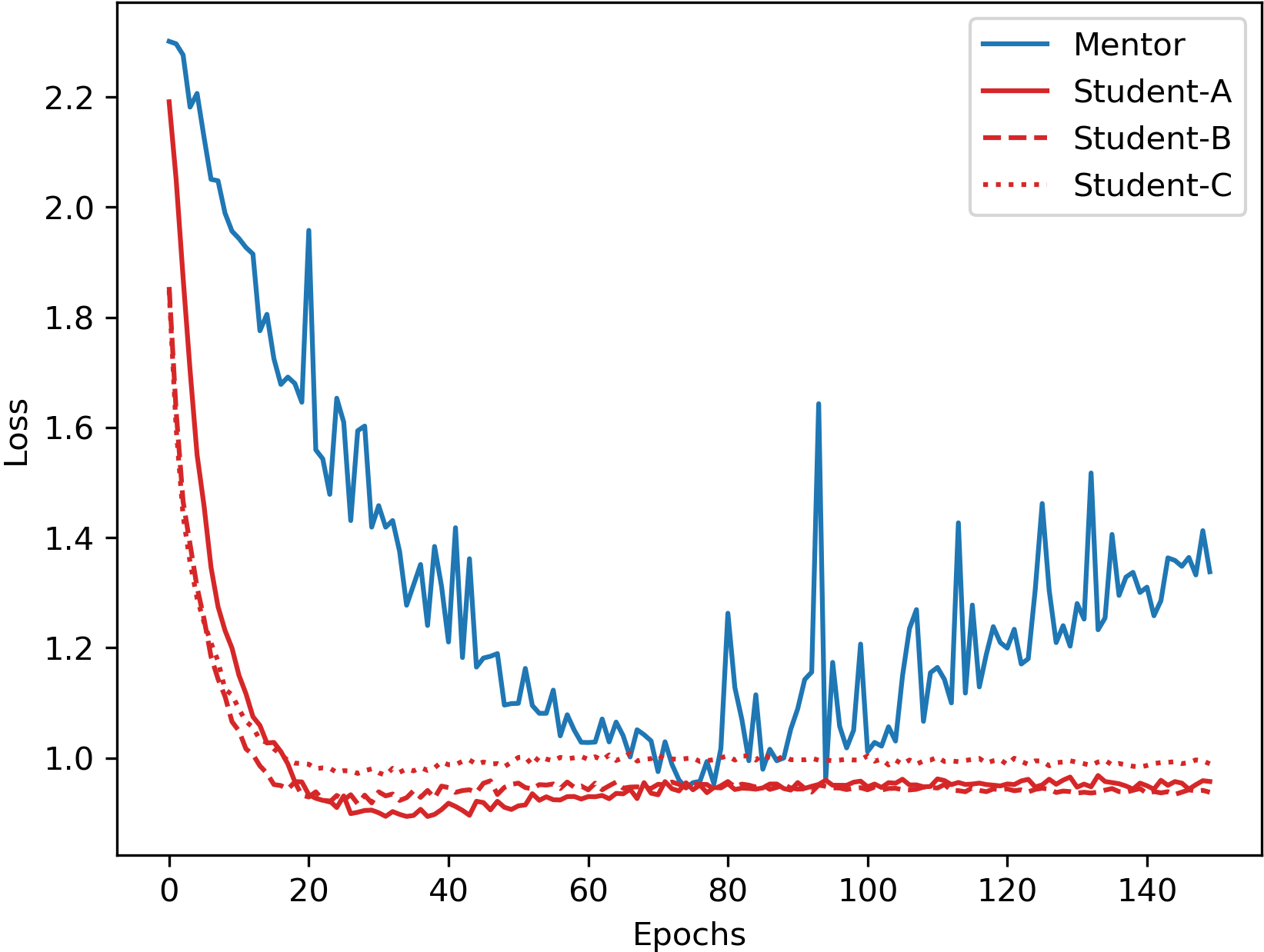}
\caption{Test Loss}
\label{sfig:testa}
\end{subfigure}%
\begin{subfigure}{.5\linewidth}
\includegraphics[width=.99\linewidth]{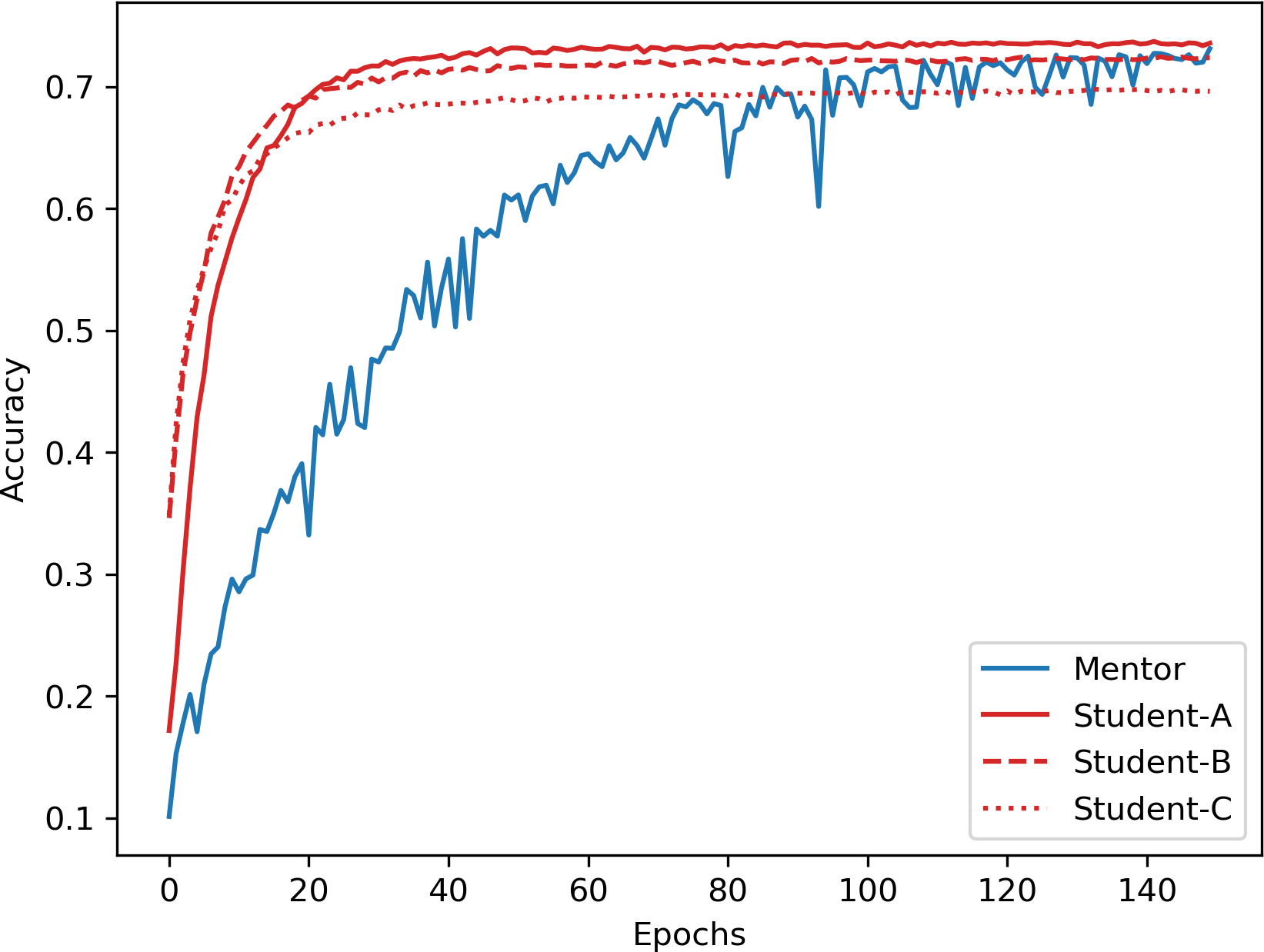}
\caption{Test Accuracy}
\label{sfig:testb}
\end{subfigure}%
\caption{Models' test loss and accuracy over 150 epochs for CIFAR-10 dataset; Students are averaged over multiple runs to show consistent results. In contrast to Mentor's spiky and increasing loss function, Student models remain steady and consistent, owing to the very strong regularization of soft label training.}
\label{cifar_acc_loss_fig}
\end{figure}

As can be seen from Table~\ref{cifar_table} and Figure~\ref{cifar_acc_loss_fig}, Student-A matches the Mentor's performance, and Student-B reaches a very high accuracy compared to that of the Mentor (only 0.76\% lower), which serves as its only training source. Finally, Student-C still reaches good results (only 3.51\% lower than the Mentor's accuracy), despite its substantially shallower architecture.

It might be of interest to observe also the training of a student model on 80\% of the data using one-hot labels instead of the mentor's predictions as a simpler mentor model training. There is a limit, of course, to simplifying the model and still obtain better accuracy than the original mentor, while training merely on 20\% of the data. In our case, Student-B and Student-C reach accuracy rates of 77.22\% and 72.64\%, respectively, while the original Mentor reaches an accuracy rate of 73.14\%. Note that the models described have four times more data to train on with simpler architectures.

\subsection{Experiments with Unbalanced CIFAR-10 Data} \label{cifar10_unbalanced_section}
\subsubsection{Reduced Student Dataset Samples}

\begin{table}[!b]
\begin{center}
\def\arraystretch{1.3}%
\begin{tabular}{cccc}
\thickhline
\setrow{\bfseries}Ratio Bound &\setrow{\bfseries} Student-A &\setrow{\bfseries} Student-B &\setrow{\bfseries} Student-C\\
\thickhline
    5\% & 73.01\%  & 71.74\% & 69.28\%\\
    10\% & 73.29\% & 71.38\% & 68.79\%\\
    20\% & 72.45\% & 71.34\% & 68.98\%\\
    30\% & 73.02\% & 71.32\% & 68.58\%\\
    40\% & 72.92\% & 70.73\% & 68.36\%\\
    50\% & 73.02\% & 70.72\% & 68.15\%\\
    60\% & 72.32\% & 69.77\% & 67.16\%\\
    70\% & 72.21\% & 69.35\% & 66.62\%\\
    80\% & 71.86\% & 69.36\% & 66.62\%\\
    90\% & 72.33\% & 69.64\% & 66.94\%\\
    
	\thickhline
\end{tabular}
\caption{Student accuracy using DeepMimic with unbalanced CIFAR-10 dataset (due to removal of data samples). Each entry is an average over multiple runs. Training is based on Mentor reaching 72.92\% accuracy on test set. All models were trained over 150 epochs.}
\label{cifar_reduction_table}
\end{center}
\end{table}

In the following experiment we tested our method on an unbalanced student dataset, as follows. After splitting the dataset by a 20\%-80\% ratio and creating a balanced student dataset, we decreased the number of samples in each class by some randomly chosen fraction to obtain an unbalanced dataset for the student. This was done on the training data alone, keeping the test set intact.
The results obtained are presented in Table~\ref{cifar_reduction_table}. We executed the experiment multiple times for different reduction bounds per each class (i.e., for different bounds on the fraction of samples removed from each class). Even for very large ratio bounds, i.e., where the amount of data available for the student model is decreased drastically, the student performance remains rather stable and the method still shows good accuracy.

\subsubsection{Added Out-of-domain Student Dataset Samples}
Having shown that an unbalanced dataset for the student model (generated by removing at random large amounts of samples from the balanced dataset) has little effect on the performance, we now demonstrate the effect of adding ``out-of-domain'' random data to the student dataset, by testing our models on this newly created dataset.
Specifically, the student dataset is modified by adding samples whose labels are very different from the categories contained in the CIFAR-10 dataset, so as to ensure non-related data to the student dataset. The labels of the added samples are, for example, Flowers, Food Containers, Fruits and Vegetables, Household Electrical Devices and Furniture, Trees, Insects, and others, taken from the CIFAR-100 dataset. As before, we use for each experiment a specified fraction limit per each class on the number of samples added at random from the other categories. The results are presented in Table~\ref{cifar_adding_table}; as can be seen, the models perform very well, reaching good accuracy with no disruption caused by the addition of out-of-domain data.

\begin{table}[!h]
\begin{center}
\def\arraystretch{1.3}%
\begin{tabular}{cccc}
\thickhline
\setrow{\bfseries}Ratio&\setrow{\bfseries} Student-A &\setrow{\bfseries} Student-B &\setrow{\bfseries} Student-C\\
\thickhline
    5\% & 73.36\% & 71.2\% & 68.96\%\\
    10\% & 73.42\%& 71.45\% & 69.04\%\\
    20\% & 73.36\%& 71.68\% & 69.01\%\\
    30\% & 73.17\%& 71.49\% & 69.35\%\\
    40\% & 73.30\%& 71.37\% & 69.05\%\\
    50\% & 73.20\%& 71.53\% & 68.96\%\\
    60\% & 73.16\%& 71.58\% & 69.04\%\\
	\thickhline
\end{tabular}
\caption{Student accuracy using DeepMimic with unbalanced CIFAR-10 dataset (due to added data samples). Each entry is an average of multiples runs. Training is based on Mentor reaching 72.92\% accuracy on test set. All models were trained over 150 epochs.}
\label{cifar_adding_table}
\end{center}
\end{table}
\subsection{Tiny ImageNet}
The Tiny ImageNet dataset is the most challenging dataset we have applied our method on.
The training data consists of 100,000 ($64 \times 64$) RGB images in 200 classes, with 500 images per class. 
There are 10,000 images in the validation set and in the test set.
As can be seen from Table~\ref{tiny_table}, the architecture used for the networks is much deeper. This makes it possible to demonstrate the effect of removing a substantial amount of layers without having almost a negative impact on the model's performance.
Note that Student-B and Student-C have much simpler architectures, yet, their obtained results are very close to the Mentor's. 

\begin{table}[!h]
\begin{center}
\def\arraystretch{1.3}%
\begin{tabular}{cccc}
\thickhline
\setrow{\bfseries}Model &\setrow{\bfseries} Architecture &\setrow{\bfseries} Accuracy &\setrow{\bfseries} Relative  Accuracy\\
\thickhline
    Mentor & $(c-bn-d)^9-fc-bn-d-s$ & 20.45\% & -\\
    Student-A & $(c-bn-d)^9-fc-bn-d-s$ & 20.47\%& 100.09\%\\
    Student-B & $(c-bn-d)^6-fc-bn-d-s$ & 20.51\% & 100.29\%\\
    Student-C & $(c-bn-d)^3-fc-bn-d-s$ & 19.60\% & 95.84\%\\
	\thickhline
\end{tabular}
\caption{Model architectures, test accuracy, and relative accuracy between Students and Mentor for Tiny ImageNet dataset. Symbols: c-convolutional layer, bn-batch norm layer, d-dropout layer, fc-fully connected layer, s-softmax layer. $\theta^n$ means $n$ consecutive layers of type $\theta$.}
\label{tiny_table}
\end{center}
\end{table}

\begin{figure}[!h]
\centering
\begin{subfigure}{.5\linewidth} %\hfill% or \hspace{5mm} or \hspace{0.3\textwidth}
\includegraphics[width=.99\linewidth]{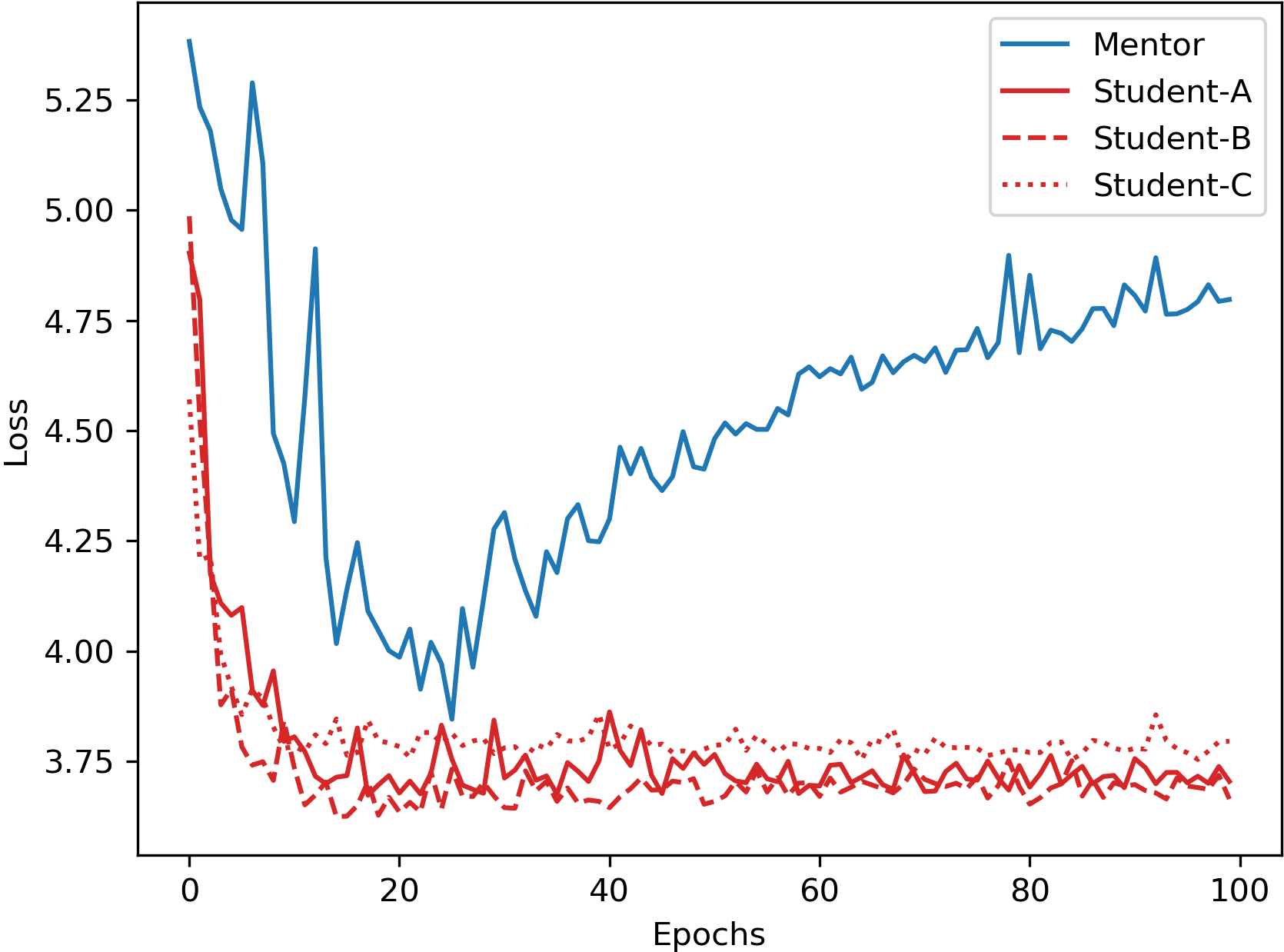}
\caption{Test Loss}
\label{sfig:testa}
\end{subfigure}%
\begin{subfigure}{.5\linewidth}
\includegraphics[width=.99\linewidth]{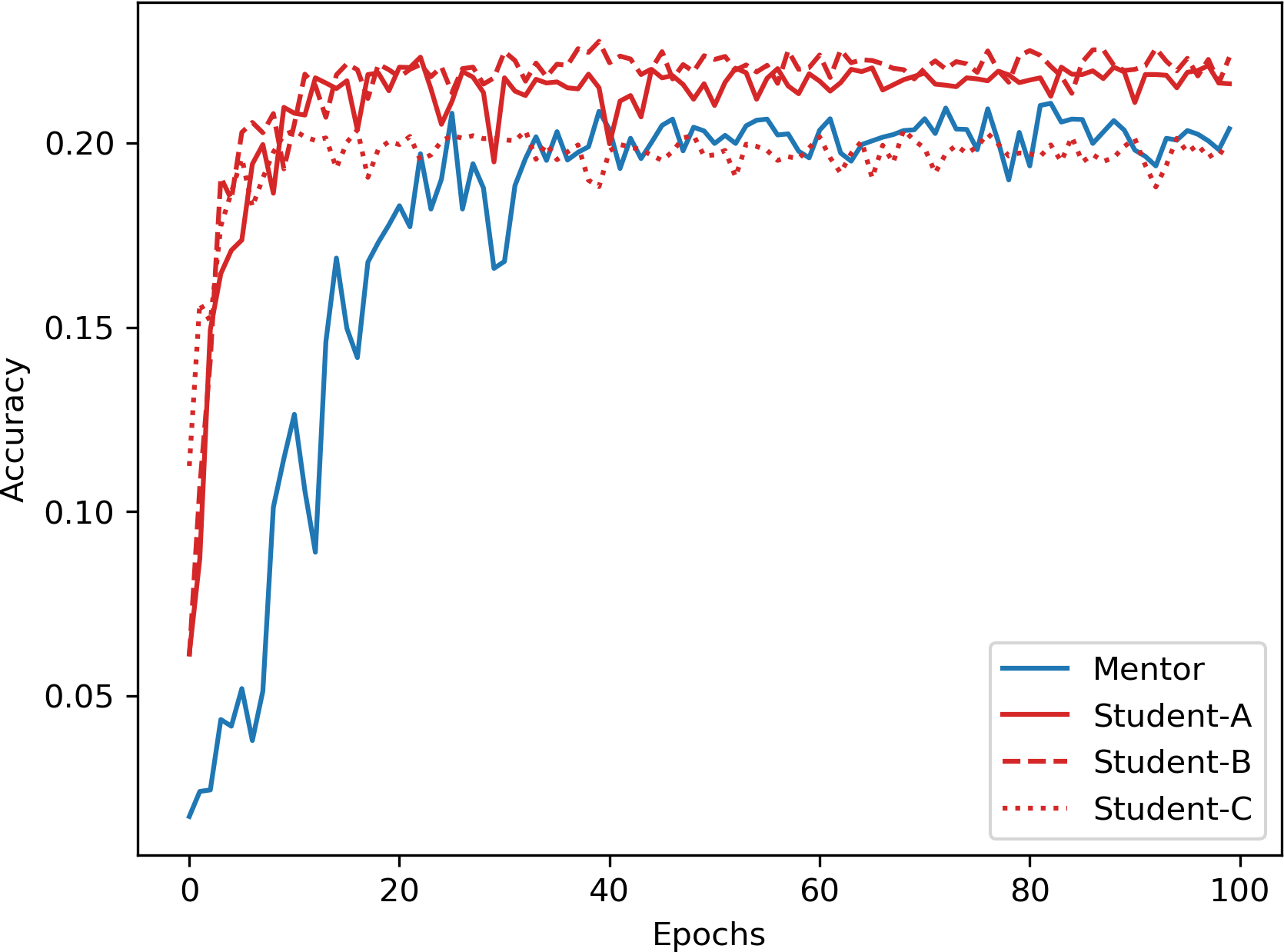}
\caption{Test Accuracy}
\label{sfig:testb}
\end{subfigure}%
\caption{Models' test loss and accuracy over 100 epochs for Tiny ImageNet dataset; Students and Mentor are averaged over multiple runs.}
\label{tiny_acc_loss_fig}
\end{figure}

\begin{figure}[!t]
\centering
\begin{subfigure}{.25\linewidth} %\hfill% or \hspace{5mm} or \hspace{0.3\textwidth}
\includegraphics[width=.9\linewidth]{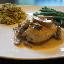}
\caption{Plate}
\label{sfig:testa}
\end{subfigure}%
\hspace{5pt}
\begin{subfigure}{.25\linewidth}
\includegraphics[width=.9\linewidth]{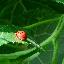}
\caption{Ladybug}
\label{sfig:testb}
\end{subfigure}%
\hspace{5pt}
\begin{subfigure}{.25\linewidth}
\includegraphics[width=.9\linewidth]{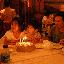}
\caption{Candle}
\label{sfig:testb}
\end{subfigure} \par\medskip
\begin{subfigure}{.25\linewidth}
\includegraphics[width=.9\linewidth]{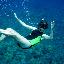}
\caption{Snorkel}
\label{sfig:testa}
\end{subfigure}%
\hspace{5pt}
\begin{subfigure}{.25\linewidth}
\includegraphics[width=.9\linewidth]{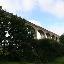}
\caption{Viaduct}
\label{sfig:testb}
\end{subfigure}%
\hspace{5pt}
\begin{subfigure}{.25\linewidth}
\includegraphics[width=.9\linewidth]{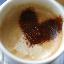}
\caption{Espresso}
\label{sfig:testb}
\end{subfigure}
\caption{Images successfully classified by both Mentor and Student-A.}
\label{fig:tiny_imagenet_correct}
\end{figure}

\begin{figure}[!h]
\centering
\begin{subfigure}{.25\linewidth}
\includegraphics[width=.9\linewidth]{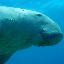}
\caption{Dugong}
\label{sfig:incorrect_a}
\end{subfigure}%
\hspace{5pt}
\begin{subfigure}{.25\linewidth}
\includegraphics[width=.9\linewidth]{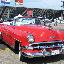}
\caption{Convertible}
\label{sfig:incorrect_b}
\end{subfigure}%
\hspace{5pt}
\begin{subfigure}{.25\linewidth}
\includegraphics[width=.9\linewidth]{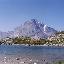}
\caption{Alp}
\label{sfig:incorrect_c}
\end{subfigure} \par\medskip
\begin{subfigure}{.25\linewidth}
\includegraphics[width=.9\linewidth]{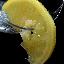}
\caption{Lemon}
\label{sfig:incorrect_d}
\end{subfigure}%
\hspace{5pt}
\begin{subfigure}{.25\linewidth}
\includegraphics[width=.9\linewidth]{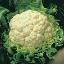}
\caption{Cauliflower}
\label{sfig:incorrect_e}
\end{subfigure}%
\hspace{5pt}
\begin{subfigure}{.25\linewidth}
\includegraphics[width=.9\linewidth]{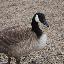}
\caption{Goose}
\label{sfig:incorrect_f}
\end{subfigure}
\caption{Images classified successfully by Mentor, but incorrectly by Student-A, classifying~\ref{sfig:incorrect_a} as Sea Cucumber, \ref{sfig:incorrect_b} as Sports Car, \ref{sfig:incorrect_c} as  Seacoast, \ref{sfig:incorrect_d} as Banana, \ref{sfig:incorrect_e} as Brain Coral, and \ref{sfig:incorrect_f} as Albatross. Although  Mentor and Student are very similar in knowledge, they are not identical.}
\label{fig:tiny_imagenet_incorrect}
\end{figure}

As can be seen in~\cite{yao2015tiny}, obtaining over 55\% accuracy on the test set is an impressive result; in contrast, a random guess yields only 0.5\% accuracy. Therefore, and considering that only a fifth of the original training data is used for training, obtaining over 20\% accuracy on the test set for the Mentor is satisfactory, as well. The result demonstrates our method's effectiveness for this dataset.
Table~\ref{tiny_table} and Figure~\ref{tiny_table} show that both Student-A and Student-B definitely match the Mentor's performance.
Student-C is the shallower model we use.
Still, it achieves only 0.85\% less accuracy than the Mentor's,  
attesting to the method's effectiveness and impressive results, even when applied to highly-complex and involved datasets. Figures~\ref{fig:tiny_imagenet_correct} 
and~\ref{fig:tiny_imagenet_incorrect} contain images classified correctly by both the Mentor and Student-A and images classified differently by the Mentor and Student-A, respectively.

\subsection{Inference Time Measurements}
\begin{table}[h]
\begin{center}
\def\arraystretch{1.5}%
\begin{tabular}{cccc}
\thickhline
\setrow{\bfseries}Dataset &\setrow{\bfseries}Model &\setrow{\bfseries} GeForce Gtx 1050 Ti &\setrow{\bfseries} GeForce Gtx 1070 \\
\thickhline
    MNIST & Mentor & 0.551  & 0.652\\
    MNIST & Student-B & 0.399  & 0.609\\
\hline
    CIFAR-10 & Mentor & 1.637  &  1.097\\
    CIFAR-10 & Student-B & 1.275  & 0.922\\
    CIFAR-10 & Student-C & 1.129 & 0.859\\
\hline
    Tiny ImageNet & Mentor & 6.328 & 3.137\\
    Tiny ImageNet & Student-B & 4.826 & 2.449\\
    Tiny ImageNet & Student-C & 4.089 & 2.194\\
	\thickhline
\end{tabular}
\caption{Inference times (in seconds) on test sets corresponding to different datasets for various models and associated mentors, using two GPU architectures. Student-A (not shown) has mentor's same architecture and hence identical speed.}

\label{inference_timings}
\end{center}
\end{table}
We have tested also comparative inference times (in seconds) for each student model versus its associated mentor, running on the test sets that correspond to the datasets experimented with (see Table~\ref{inference_timings}).
Each model was tested on two different GPU architectures,
with the results averaged over 100 executions. When using a more complex and deeper network, which is usually the case in real-life scenarios, the time reduction is more significant, and may allow for much faster data processing. Sometimes the student seems to slightly surpasses the mentor; this behavior was observed mostly for student models which are replicas of the mentor,
or a student with relatively little reduction in architecture. 
Determining whether a smaller, albeit less accurate model, should be used versus a larger, more accurate model, is an interesting question. For DNN-based cloud services, the answer would probably be never, as such services usually rely on very strong and expensive hardware, so we would not be limited by any restrictions and just use the most accurate model.
However, embedded devices which usually do not rely on strong hardware or stable internet connection, e.g., a cell phone or an IOT (Internet of things) device, are mostly more limited as far as size, memory, and power.
The manufacturers would usually develop an extremely small and less powerful hardware, in order to keep the product small, elegant, and rather inexpensive. 
The mentioned limitations are quite problematic when one is interested in deploying a massive model on a product.
In such scenarios, creating a significantly smaller and faster model would enable to deploy it on a smaller hardware, so it is highly likely that manufacturers 
would rather employ a less accurate model than a more accurate one which cannot be embedded in their products.

\section{Conclusion}
In this paper, we have presented a novel approach for training deep neural networks.
Our DeepMimic method relies on utilizing two models, which are not necessarily identical.
We have shown that reducing the student model's complexity has a minor effect on its success rate compared to the mentor's. According to this empirical evidence, it is possible to mimic a black-box mentor model with an unknown architecture and reach the same accuracy.
In a series of experiments, we have shown that for both balanced and unbalanced training data available for the student, the method manages to mimic the mentor model successfully.
One only needs to exploit large amounts of unlabeled data, which is the expected scenario in real-life situations.
Our method raises serious security implications, as one can ``duplicate'' a proprietary neural network, by creating a copy of it without having access to the original training data.
The method presented yields impressive results and exploits large amounts of unlabeled data for training, without having to manually tag them. We have worked solely on CNNs for both the mentor and student models. Our method can be further extended and used to explore the relations between different types of networks, e.g., a fully-connected network and a CNN.

This could prove as a key factor to obtain, extract, and transfer knowledge between different types of networks, thereby pushing further the performance level.

\section{Future Work}
As can be seen from Table~\ref{tiny_table}, the larger the network, the easier it is to reduce its size more significantly with low reduction in accuracy. In such cases, the effect on the inference time is more noticeable and such compressed networks 
have an advantage, as shown in
Table~\ref{inference_timings}. Therefore, we would prefer to test our method on deeper networks such as VGG~\cite{deep_convolutional_2014} and ResNet~\cite{2016_risidual}, expecting to create models with even more improved inference times. So far we have experimented mainly with CNNs for classification problems, but it is of interest to explore the effect of DeepMimic in other problem domains, e.g., networks designed for detection and segmentation. Such networks usually perform feature extraction on the input and rely on massive architectures to do so, we expect our method to be very beneficial in these domains.

An additional idea that might lead to a much smaller, yet a more accurate student, is to distill multiple mentor models into a single student model. By doing so, the student training data can be increased by using multiple mentors to generate the data
or we could average different mentor predictions to make the student hopefully more accurate. 

An interesting work regarding CNN classifiers using low-shot learning is given in~\cite{lowshot2018}. The idea is to enable a model to successfully classify a newly seen category after being presented with merely few training examples. This notion resembles the way human vision works using imprinted weights. The authors use a CNN as an embedding extractor, and after a classifier is trained, the embedding vectors of new low-shot examples are used to imprint weights for new classes in the extended classifier. As a result, the new model is able to classify well examples belonging to a novel category after seeing only a few examples.
Combining this work and DeepMimic might be very interesting, in the following sense.
While using a mentor model trained on specific categories, upon the arrival of a novel category it might be easier to implant the new category in a student model combining the two processes described in DeepMimic and~\cite{lowshot2018}. It is possible that a student model would adjust more naturally to new categories during the training process itself rather 
than an already trained model.

\bibliographystyle{splncs04}
\bibliography{deepmimic}
\end{document}